\documentclass[runningheads]{llncs}
\usepackage[T1]{fontenc}
\usepackage{graphicx}
\begin{document}
\title{A Simple Structure For Building A Robust Model\thanks{Supported by AutoDL.}}
%
%
\author{Xiao Tan\inst{1,2}\orcidID{+8617610072986} \and
JingBo Gao\inst{1}\orcidID{+8618667090860} \and
Ruolin Li\inst{1}\orcidID{+8615399311502}}
\authorrunning{ }
%
\institute{
Xidian Hangzhou Institute of Technology, Zhejiang Province, China
\url{https://hz.xidian.edu.cn} \and
\email{1203550038@qq.com}
}
\maketitle              
\begin{abstract}
As deep learning applications, especially programs of computer vision, are increasingly deployed in our lives, we have to think more urgently about the security of these applications.One effective way to improve the security of deep learning models is to perform adversarial training, which allows the model to be compatible with samples that are deliberately created for use in attacking the model.Based on this, we propose a simple architecture to build a model with a certain degree of robustness, which improves the robustness of the trained network by adding an adversarial sample detection network for cooperative training.At the same time, we design a new data sampling strategy that incorporates multiple existing attacks, allowing the model to adapt to many different adversarial attacks with a single training.We conducted some experiments to test the effectiveness of this design based on Cifar10 dataset, and the results indicate that it has some degree of positive effect on the robustness of the model.Our code could be found at \url{https://github.com/dowdyboy/simple\_structure\_for\_robust\_model}.

\keywords{Robustness  \and Adversarial Training \and Deep Learning}
\end{abstract}
\section{Introduction}

Currently, applications using deep learning methods are gradually and widely used in our lives\cite{bib_1,bib_2,bib_3}.In many scenarios, deep learning algorithms and models must be secure.However, models trained by general deep learning methods are often very sensitive to the input data.Small changes in the input data can lead to large deviations in the model output, which makes it possible to spoof the model by falsifying data that is indistinguishable to the human eye.Such artificially created samples used to deceive the model are called adversarial samples\cite{bib_4}.

To reduce the negative impact of such problems, there are two mainstream approaches, one is adversarial sample detection\cite{bib_5} and the other is training robust models\cite{bib_6}.Both schemes have their advantages and disadvantages, and both can mitigate the impact of attacks on the model to some extent. In practical application scenarios, the two approaches often need to be used in conjunction\cite{bib_7}, as many adversarial samples are difficult to detect or have a high error detection rate.

\begin{figure}
	\includegraphics[width=\textwidth]{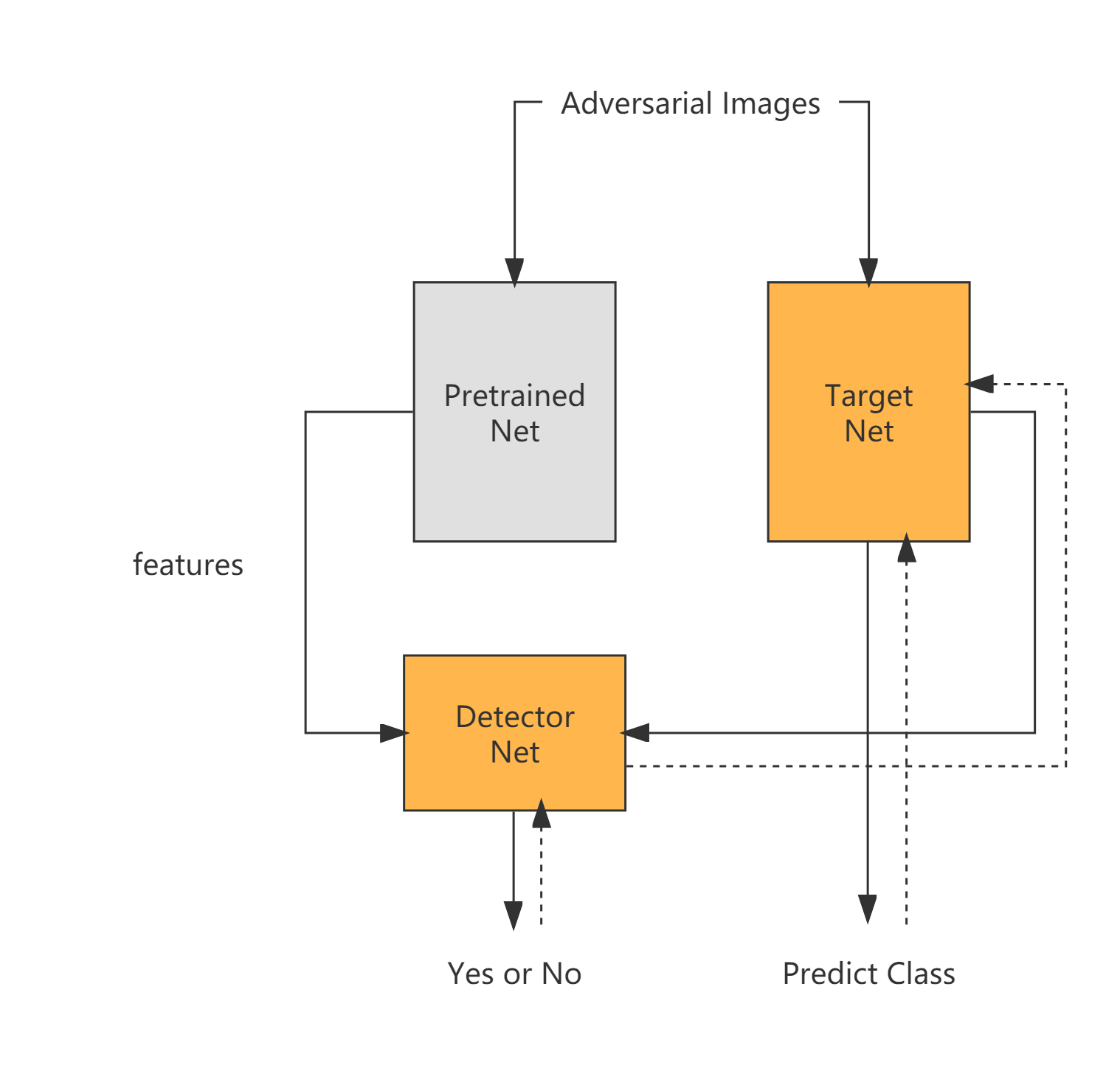}
	\caption{Pretrained Net and Target Net have the same network structure. The input samples are entered into these two networks separately and the intermediate feature maps of the network outputs are obtained. The two feature maps compute the differences in Detector Net and provide feedback and fine-tuning to the upstream network.} \label{fig_model_design}
\end{figure}

We carefully analyzed the existing methods\cite{bib_8,bib_9} and found that the two solutions can be used not only in combination at the application level, but also in conjunction with each other at the training stage.Therefore, we propose a simple structure(Fig.~\ref{fig_model_design}), which improves the efficiency of robustness training through the intermediate results of adversarial sample detection.This architecture has excellent scalability, as it is suitable for almost any network structure.With the control backbone network unchanged, we also designed different adversarial sample detection networks and conducted experiments, and found some interesting results(Table.~\ref{tab_in_compare}).

In a real deployment environment, models need to face a wide variety of attacks\cite{bib_10}.After some investigation, we found that the data sampling strategy has an impact on the training of the model\cite{bib_11,bib_12}.Based on this, we propose a training data sampling strategy which can obtain a richer sample type with the same training data size.This approach allows the model to be adapted to more complex test scenarios in a single training session.In addition, we adopt an offline data processing scheme\cite{bib_13}, which requires more storage space but can effectively improve the training speed.

We tested our designed structure using the Cifar10 dataset\cite{bib_14} and found that our scheme can improve the robustness of the model to some extent compared to the common adversarial training method\cite{bib_6,bib_15}.Also, because of the new data sampling strategy, the model is able to be more adaptable to more complex testing environments.

\section{Related Work}
In recent years, the problem of robustness of deep learning models has attracted increasing attention, and numerous attack methods and defense methods have been proposed.Here, we will discuss those methods that are relevant to our schema and what they bring to our design.

\subsection{Adversarial Attacks}
Deep learning has developed rapidly and has been widely used in many areas, such as speech recognition\cite{bib_16}, image classification\cite{bib_17}, and object detection\cite{bib_18}. However, it also comes with the problem of its security. In 2013, Szegedy et al. found that adding small perturbations to images can trick deep neural networks (DNN) with high probability\cite{bib_19} into producing incorrect classification results, and these misclassified samples are called adversarial samples.

\subsubsection{Gaussian Noise} The simplest adversarial attack is to add Gaussian noise to the image\cite{bib_20}.These Gaussian noises, because they possess local randomness, can interfere with the pixel value distribution of the image.Although these images with simple Gaussian noise added do not produce large changes compared to the original (and are not misleading to the human eye), they can still lead to discrimination errors in simple models.

\subsubsection{FGSM} FGSM was proposed by Goodfellow et al. and is a classical algorithm in the field of adversarial samples\cite{bib_21}.The main reason for the vulnerability of neural networks to adversarial perturbations is their linear nature and, the linear behavior in high-dimensional spaces is sufficient to cause misclassification of samples.FGSM is based on this principle and generates adversarial samples with the help of gradients and the goal of maximizing the loss function\cite{bib_22}.

\subsubsection{BIM} The BIM algorithm\cite{bib_23} uses an iterative approach to search for perturbations at individual pixel points, rather than modifying all pixel points at once as a whole.During the iterative update process, some pixel values of the sample may overflow as the number of iterations increases.These values need to be replaced by 0 or 1 in order to finally generate a valid image, which is used to ensure that each pixel of the new sample is within a certain field of each pixel of the original sample.

\subsubsection{DeepFool} The DeepFool algorithm\cite{bib_24} adds less noise and takes less time to generate samples compared to the FGSM algorithm.It is based on the classification idea of hyperplane and can accurately calculate the perturbation value.Adversarial training using DeepFool samples can effectively improve the robustness of the model.

\subsubsection{C\&W} The C\&W algorithm\cite{bib_25} is an optimization-based attack algorithm.It sets a special loss function to measure the difference between the input and the output.This loss function contains adjustable hyperparameters, as well as parameters that control the confidence level of the generated adversarial samples.By choosing appropriate values for these two parameters, excellent adversarial samples are generated.

\subsubsection{NST} Neural Style Transfer is a way to perform style changes on images\cite{bib_26}.Since the texture of the image is modified during the transformation of the image, it can obviously be applied to generate adversarial samples as well.These adversarial samples deceive the model in terms of the underlying texture and direct the model output to the wrong class.

\subsection{Adversarial Defensive}
Currently popular adversarial defense methods include model distillation\cite{bib_27} and adversarial training\cite{bib_6}.Model distillation uses a teacher model to guide the training of the student model, and the teacher model is a pre-trained model.When training is performed, the input data are first entered into the teacher model and the probability distribution of the output is obtained.Training the student model with this probability distribution and the original input data allows the student model to quickly learn the capabilities of the teacher model.Also, Since the probability distribution of the teacher model output is used as the training label, it can hide the gradient information of the model to some extent\cite{bib_28}, thus making the trained student model robust to some adversarial attacks, but it is not effective for attack methods like C\&W\cite{bib_25}.

Adversarial training is a commonly used method to improve the robustness of models.It is to exploit the powerful expressive power of deep neural networks\cite{bib_29} to improve the robustness of the model by learning adversarial samples.On top of classical adversarial training, it can also incorporate self-supervised tasks to obtain pre-trained models\cite{bib_30} and can be fine-tuned using expanded datasets.These techniques can further strengthen the model.Adversarial training also has certain limitations, such as the possibility of overfitting the employed attack method and the adversarial sample\cite{bib_31}.To address these issues, our approach uses a detection head for assisted training as a way to further strengthen the effectiveness of the model, and on the other hand, it uses a data sampling strategy that can combine more attack methods and adversarial samples to prevent overfitting.

\section{Methods}
\subsection{Structure Design}
Adversarial training is an effective way to resist adversarial samples and improve model robustness.However, due to the large number of adversarial samples added to the training data, this leads to a longer convergence process of the training.Inspired by the design of the auxiliary head\cite{bib_32}, our method also adopts a similar design as a way to improve the model convergence speed.Also, in the selection of the auxiliary head, instead of using the auxiliary head of the same type of task, we use the adversarial sample detection task as the auxiliary head training method because we believe that intuitively, the adversarial sample detection task and the adversarial training have some correlation and can extract different characteristics from the adversarial sample.

\begin{figure}
	\includegraphics[width=\textwidth]{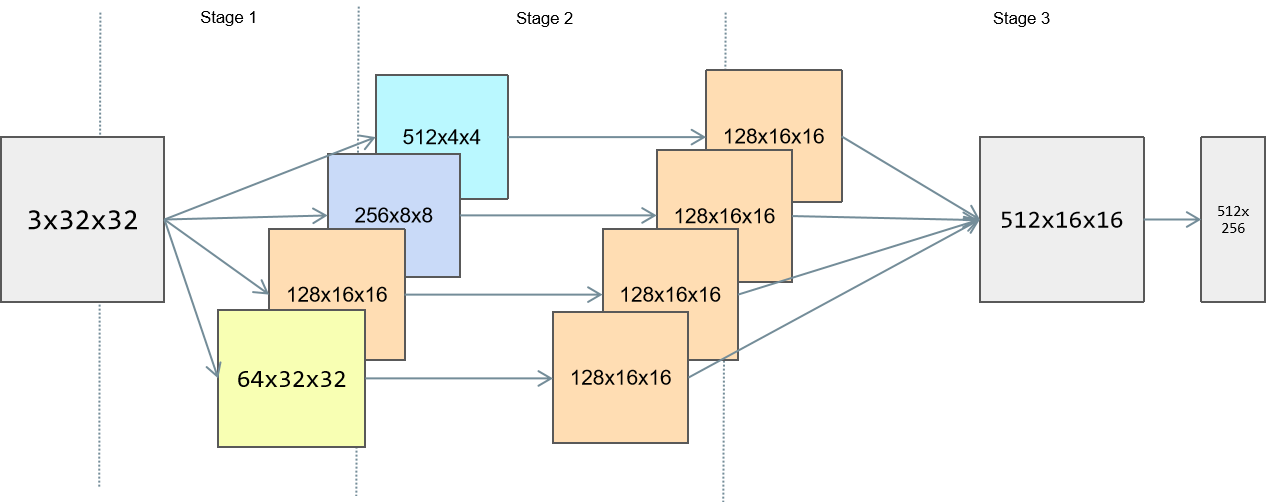}
	\caption{The adversarial detection network receives the image input and goes through three stages of processing: in the first stage, the data is fed into the backbone network to obtain feature maps of different layers; in the second stage, data embedding is performed, and the feature maps are convolved or deconvoluted and outputted uniformly as feature maps of the same size and number of channels; in the third stage, all the feature maps are stitched and spread into the output that Transformer can receive.} \label{fig_embed}
\end{figure}

We designed a simple shallow Transformer\cite{bib_33} structure as an auxiliary network to perform the adversarial sample detection task.However, unlike a normal network model, its input is not the original data, but a feature map of the output of an intermediate layer of a network that has been pre-trained on a clean dataset.And, inspired by Vision in Transformer\cite{bib_34}, we abandon the use of fully connected layers as embedding layers and use convolutional structures as embedding layers instead(Fig.~\ref{fig_embed}).Since our input needs to fuse feature maps from different levels of the pre-trained network as input, we introduce deconvolution in our embedding layer\cite{bib_35}.After experiments, we found that fusing other levels of feature maps with a shallow feature map as the center can have better results.

Because the secondary and primary networks perform different classes of tasks, they cannot simply perform gradient returns and updates separately.For this case, our approach is to capture the output of a specific layer in the auxiliary network and then find the differential performance of the output features of the pre-trained network and the target network on the auxiliary network by using the Smooth L1 function\cite{bib_22} as part of the loss function.The final loss function consists of the categorical cross-entropy error and the L1 error [equation \ref{eq_1}].
\begin{equation}
	 loss(X,Y,L) = \sum_{i=1}^{N} \alpha \times CrossEntropyLoss(X_{i},L_{i}) + \beta \times SmoothL1Loss(X_{i},Y_{i})  \label{eq_1}
\end{equation}
\subsection{Sampling Strategy}
Classical sampling strategies for adversarial training tend to use a fixed ratio of adversarial samples as input\cite{bib_6}, however, this approach may in some cases make the network overfit to the selected adversarial sample data\cite{bib_31}, because the type of adversarial attack to which a particular picture belongs under that sampling strategy is specific to a particular one.Faced with this situation, we designed a more flexible sampling strategy.It uses a dynamic probability with a bounded range to determine the type of adversarial attack to which the sample belongs.Due to the random nature of dynamic probabilities, each image generates more types of adversarial samples, thus enriching the entire training dataset.

\begin{figure}
	\includegraphics[width=\textwidth]{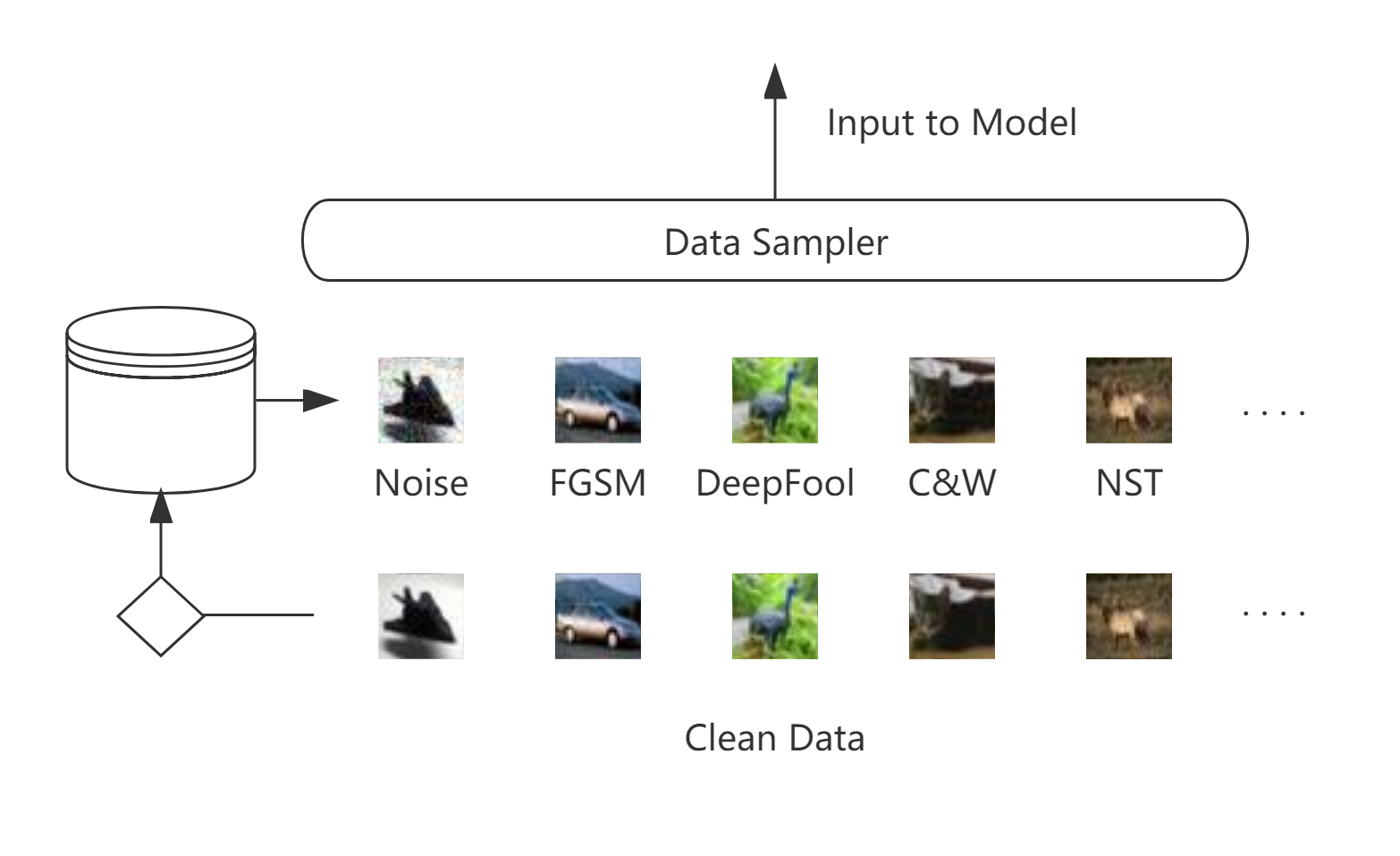}
	\caption{The Clean dataset is programmed in an offline manner to generate adversarial samples of all categories and stored in a database. The sampler samples the database according to the configured dynamic probability range and feeds the selected samples to the model.} \label{fig_data_sampler}
\end{figure}

In addition, unlike the scheme of generating adversarial samples in the process of training, we take the approach of generating adversarial samples offline (Fig.~\ref{fig_data_sampler}).Adversarial samples of all attack types are generated for each sample in the training dataset and stored in the database before starting training.To perform training, adversarial samples of the specified type are selected from the database as input to the model based on dynamic probabilities.Offline generation of adversarial samples consumes more storage space, but can improve the efficiency of the training process.It also does not reduce the diversity of the sample space due to the use of a sampling strategy with greater randomness.

\section{Experiments}
For our proposed model architecture, we conducted a series of experiments.All of our experiments were run on a dual-card RTX A4000 mainframe.The experimental results show that our method is able to make the network model more robust compared to ordinary adversarial training(Table.~\ref{tab_base}).In addition, we also tried some detailed modifications, including the proportion of adversarial samples, the structure of the detection head, and the weight of the loss function, and found some laws that affect the performance of the model(Table.~\ref{tab_in_compare}).

\begin{table}
	\centering
	\caption{Intensity of sample noise under different types of adversarial attacks.}\label{tab_nosie_cfg}
	\begin{tabular}{|l|l|l|l|l|l|l|}
		\hline
		Name: &  Gaussian & FGSM & BIM & DeepFool & C\&W & NST \\
		\hline
		Value: & 0.1 & 0.005 & 0.001 & 0.1 & 0.005 & 0.1 \\
		\hline
	\end{tabular}
\end{table}

We used the Cifar10 dataset\cite{bib_14} for our experiments.We randomly sampled 45,000 images from the original training set as our training set and the remaining 5,000 images as the test set.The reason for this is that we want to keep the data distribution of the training and test sets as consistent as possible.We generated datasets with adversarial sample ratios of 0.5, 0.75, and 0.25, and the types of adversarial samples contain FGSM, DeepFool, etc., and they use the appropriate noise intensity(Table.~\ref{tab_nosie_cfg}), respectively.

\begin{table}
	\centering
	\caption{Comparison of the prediction results of the ordinary model, the adversarial training model and our method, in terms of accuracy.}\label{tab_base}
	\begin{tabular}{|l|l|l|}
		\hline
		Structure &  Adv. Data Ratio & Accuracy \\
		\hline
		Baseline & 0.5 & 0.6778 \\
		Adv Train & 0.5 & 0.8464 \\
		{\bfseries Adv Train + Detector (ours)} & 0.5 & {\bfseries 0.8688} \\
		\hline
	\end{tabular}
\end{table}

We chose ResNet34\cite{bib_36} as our experimental network, and we modified the network model so that it can output all intermediate feature maps.We first performed training on a clean dataset and general adversarial training, after which we conducted experiments using our method on the same dataset(Table.~\ref{tab_nosie_cfg}), and finally, we conducted some exploratory experiments for different parameter configurations of our method(Table.~\ref{tab_in_compare}).In addition, experiments were conducted on a single adversarial sample test set in order to more fully validate the expressiveness of the model(Table.~\ref{tab_in_detail}).The number of iterations in the training phase for all models is 150 epochs.

\begin{table}
	\centering
	\caption{Comparison of results on adversarial sample detection and classification for different input types, adversarial sample ratios, and number of self-attentive layers.}\label{tab_in_compare}
	\begin{tabular}{|l|l|l|l|l|}
		\hline
		Input Type &  Adv. Data Ratio & SA Layers & Detector Accuracy & Classification Accuracy \\
		\hline
		Low & 0.5 & 4 & 0.734 & 0.8658 \\
		Low & 0.75 & 4 & 0.8332 & 0.842 \\
		Low & 0.75 & 8 & 0.8302 & 0.845 \\
		Low & 0.5 & 2 & 0.735 & 0.8694 \\
		Low & 0.75 & 2 & 0.8364 & 0.8416 \\
		\hline
		Full & 0.5 & 4 & 0.7538 & 0.8674 \\
		Full & 0.75 & 4 & 0.8458 & 0.8406 \\
		Full & 0.75 & 8 & 0.8492 & 0.8446 \\
		Full & 0.5 & 2 & 0.7506 & 0.8688 \\
		Full & 0.75 & 2 & 0.8536 & 0.8462 \\
		\hline
	\end{tabular}
\end{table}

With the same adversarial sample ratio, our method has some improvement in accuracy over the general adversarial training(Table.~\ref{tab_base}).Also, our method is more robust to each different type of adversarial attack(Table.~\ref{tab_in_detail}), which indicates that it does not have a significant bias on the performance improvement and is a more general method to improve the robustness of the model.

We also tried to configure different input types, adversarial sample ratios, and the number of self-attentive modules for the detection network(Table.~\ref{tab_in_compare}).We find that embedding all level feature maps can improve the accuracy of the detection network compared to embedding only low level feature maps, but has no significant effect on the classification accuracy of the images.In addition, excessively increasing the ratio of adversarial samples may not improve the performance of the model, as too many adversarial samples may make it more difficult for the model to capture the essential features needed to perform the classification.And, adding more self-attentive modules does not improve the performance of the model, but leads to a small decrease, which may be due to the fact that for capturing detectability features with neural networks, too many layers are not needed, otherwise some useless information may be captured, thus causing negative feedback to the target network.

\begin{table}
	\centering
	\caption{Prediction results of different structures on different types of adversarial sample datasets.}\label{tab_in_detail}
	\begin{tabular}{|l|l|l|l|l|l|l|l|}
		\hline
		Structure &  Clean & Gaussian & FGSM & BIM & DeepFool & C\&W & NST \\
		\hline
		Baseline & 0.9012 & 0.6616 & 0.307 & 0.1748 & 0.1984 & 0.1074 & 0.6236 \\
		Adv Train & 0.8737 & 0.8282 & 0.751 & 0.7906 & 0.867 & 0.7736 & 0.836 \\
		{\bfseries Adv Train + Detector (ours)} & {\bfseries 0.8916} & {\bfseries 0.8542} & {\bfseries 0.7834} & {\bfseries 0.8296} & {\bfseries 0.8884} & {\bfseries 0.8066} & {\bfseries 0.8592} \\
		\hline
	\end{tabular}
\end{table}

\section{Conclusion}
As the security and robustness of deep learning models are increasingly emphasized, adversarial attacks may gradually become a necessary factor to be considered for training models.An effective means to deal with adversarial attacks is to train the model adversarially, which enhances the performance and robustness of the model in a hostile environment by strengthening its adaptation to adversarial samples.Our proposed method adds a small adversarial sample detection network to the adversarial training by which some additional features of the data are extracted and used to improve the efficiency and performance of the adversarial training.In addition, we design a new offline data sampling strategy that incorporates a variety of adversarial attack types and has stronger randomness, allowing the trained model to be more adaptable to complex environments while improving the training efficiency to a certain extent.After experiments, we found that the above method has certain effectiveness in practice.Also, since this design structure is very expandable, optimizing some design and configuration details on the basis of this structure may further increase the model's capability.

\end{document}